\documentclass[10pt,twocolumn,letterpaper]{article}

\usepackage{wacv}              

\usepackage{times}
\usepackage{epsfig}
\usepackage{graphicx}
\usepackage{amsmath}
\usepackage{amssymb}
\usepackage{pifont}
\newcommand{\cmark}{\ding{51}}%
\newcommand{\xmark}{\ding{55}}%
\usepackage[table,xcdraw,dvipsnames]{xcolor}
\usepackage[LGRgreek]{mathastext}
\usepackage{booktabs}

\usepackage{hyperref}
\hypersetup{breaklinks=true, colorlinks=true}

\usepackage[capitalize]{cleveref}
\crefname{section}{Sec.}{Secs.}
\Crefname{section}{Section}{Sections}
\Crefname{table}{Table}{Tables}
\crefname{table}{Tab.}{Tabs.}

\def\wacvPaperID{1092} 
\def\confName{WACV}
\def\confYear{2024}

\definecolor{fig2gray}{RGB}{189, 199, 211}
\definecolor{fig2pink}{RGB}{245, 219, 214}
\definecolor{fig2blue}{RGB}{209, 227, 253}
\definecolor{fig2green}{RGB}{217, 231, 214}

\begin{document}

\title{MaskConver: Revisiting Pure Convolution Model for Panoptic Segmentation}

\author{
Abdullah Rashwan\\
Google\\
{\tt\small arashwan@google.com}
\and
Jiageng Zhang\\
Google\\
{\tt\small jiageng@google.com}
\and
Ali Taalimi\\
Google\\
{\tt\small taalimi@google.com}
\and
Fan Yang\\
Google\\
{\tt\small fyangf@google.com}
\and
Xingyi Zhou\\
Google\\
{\tt\small zhouxy@google.com}
\and
Chaochao Yan\\
Google\\
{\tt\small allenyan@google.com}
\and
Liang-Chieh Chen\\
ByteDance \thanks{Work done while at Google.}\\
{\tt\small lcchen@cs.ucla.edu}
\and
Yeqing Li\\
Google\\
{\tt\small yeqing@google.com}
}

\maketitle

\begin{abstract}

In recent years, transformer-based models have dominated panoptic segmentation, thanks to their strong modeling capabilities and their unified representation for both semantic and instance classes as global binary masks.
In this paper, we revisit pure convolution model and propose a novel panoptic architecture named MaskConver. MaskConver proposes to fully unify things and stuff representation by predicting their centers. To that extent, it creates a lightweight class embedding module that can break the ties when multiple centers co-exist in the same location. 
Furthermore, our study shows that the decoder design is critical in ensuring that the model has sufficient context for accurate detection and segmentation. We introduce a powerful ConvNeXt-UNet decoder that closes the performance gap between convolution- and transformer-based models.
With ResNet50 backbone, our MaskConver achieves 53.6\% PQ on the COCO panoptic val set, outperforming the modern convolution-based model, Panoptic FCN, by 9.3\% as well as transformer-based models such as Mask2Former (+1.7\% PQ) and kMaX-DeepLab (+0.6\% PQ).
Additionally, MaskConver with a MobileNet backbone reaches 37.2\% PQ, improving over Panoptic-DeepLab by +6.4\% under the same FLOPs/latency constraints.
A further optimized version of MaskConver achieves 29.7\% PQ, while running in real-time on mobile devices. The code and model weights will be publicly available. \footnote{ \url{https://github.com/tensorflow/models/tree/master/official/projects/maskconver}.}

\end{abstract}
\section{Introduction}

\begin{figure}[ht]
\begin{center}
\includegraphics[width=1.0\linewidth]{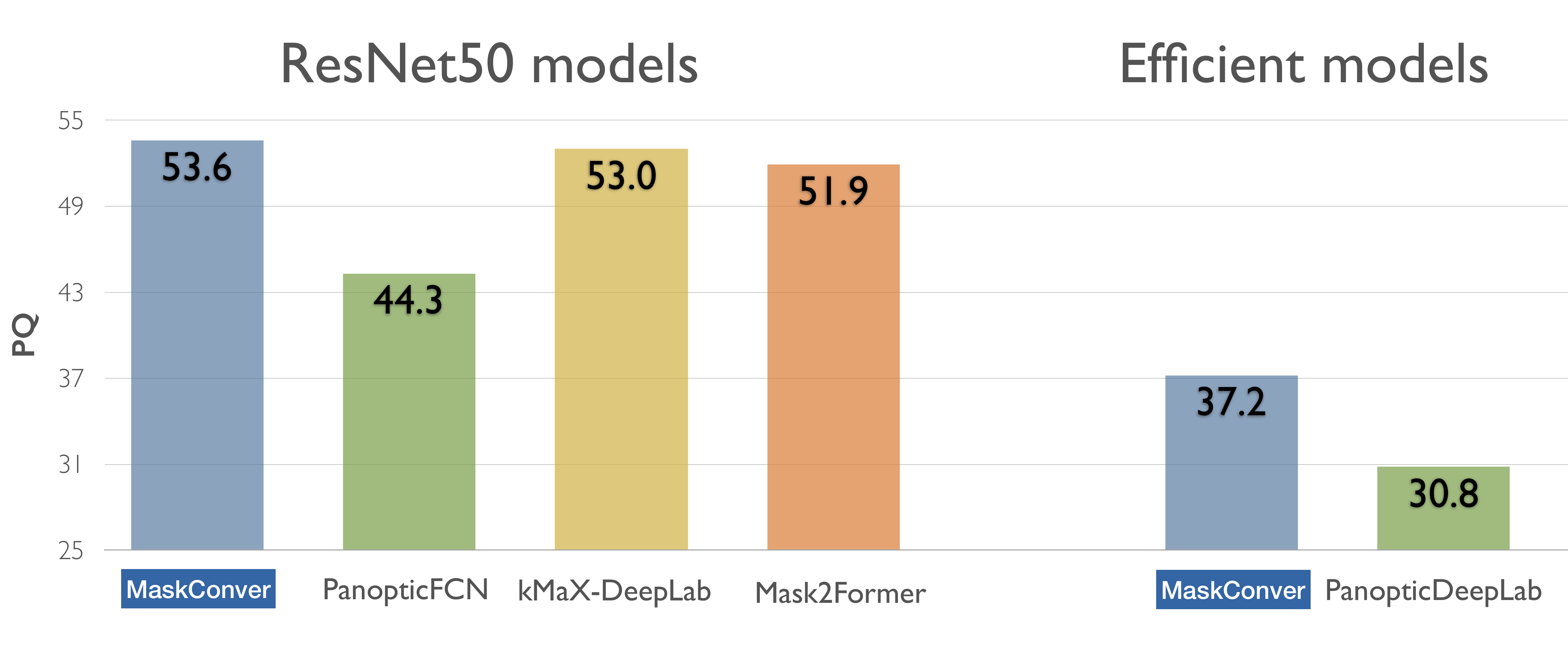}
\end{center}
   \caption{
   \textbf{MaskConver versus existing panoptic models.} MaskConver successfully bridges the gap between the modern convolution-based method, Panoptic FCN, and the transformer-based methods, Mask2Former and kMaX-DeepLab.
   In the efficient model setting, MaskConver outperforms Panoptic-DeepLab under the same FLOPs/latency constraints.
  }
\label{fig:teaser}
\vspace{-0.3cm}
\end{figure}

Panoptic segmentation~\cite{kirillov2019panoptic} aims to unify instance~\cite{hariharan2014simultaneous} and semantic segmentation~\cite{he2004multiscale} in the same framework.
Existing works propose
to merge instance and semantic segmentation outputs using post-processing layers~\cite{kirillov2019panopticfpn,xiong2019upsnet,liu2019end,yang2019deeperlab}.
These architectures however rely on many customized components like non-maximum suppression (NMS), and thing-stuff merging heuristics to produce panoptic outputs. 
Recent works~\cite{wang2021max,cheng2021per,cheng2022masked,li2022panoptic,yu2022cmt,yu2022kmax} unify both segmentation tasks by producing binary masks and class scores for both things and stuff classes. 
Such universal architectures result in a simpler post-processing logic and make the loss closely correlated with the panoptic quality (PQ) metric. 
As a result, they have achieved significantly higher PQ numbers compared to traditional architectures.

Among these unified panoptic segmentation models, transformers~\cite{vaswani2017attention} have played a critical role due to their ability to learn instance-level embeddings via a transformer decoder.
DETR~\cite{carion2020end} was introduced for object detection by learning object embeddings, each of which predicts an object class and a bounding box.
The idea of instance-level embeddings influenced many of the transformer-based panoptic models.
Following DETR, MaX-DeepLab~\cite{wang2021max} uses a transformer decoder to learn mask embeddings~\cite{jia2016dynamic,tian2020conditional,wang2020solov2} to predict a set of binary masks. The binary masks are then merged using a simple post-processing layer~\cite{mark2021deeplab2} to filter out duplicates, similar to NMS.
Other architectures~\cite{cheng2021per,cheng2022masked,li2022panoptic,yu2022cmt,yu2022kmax} follow a similar paradigm, but further improved the performance by developing modern transformer decoders~\cite{zhu2021deformable}.
The commonality between these methods is the employment of transformer blocks to learn a set of binary masks and their corresponding classes (hence the panoptic masks).
On the other hand, the convolution-based methods~\cite{cheng2020panoptic,li2021fully} lag behind in performance.
It is yet unclear if using transformers justifies the quality gap compared to convolution-based models.

\begin{figure*}[ht]
\begin{center}
\includegraphics[width=0.80\linewidth]{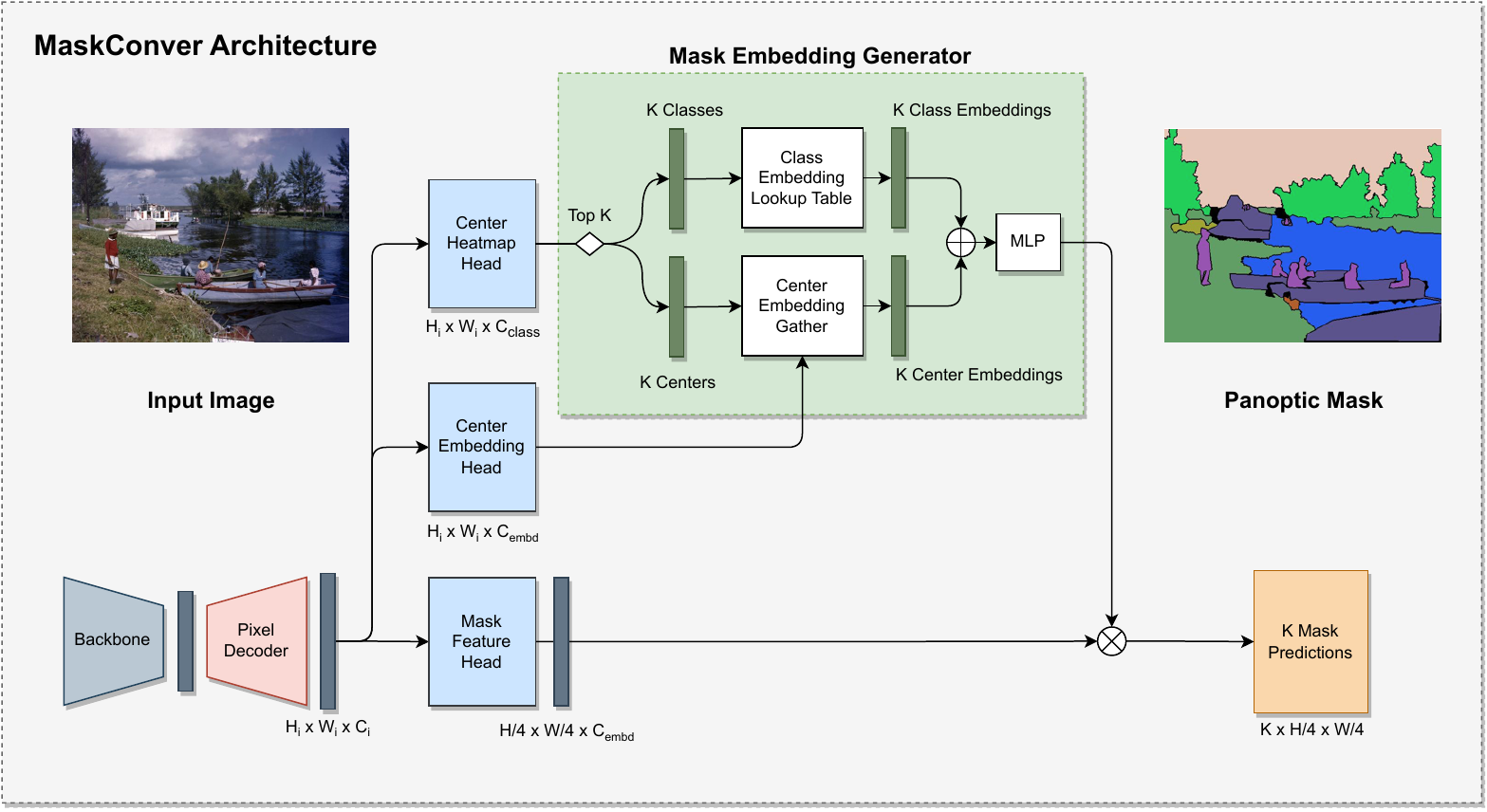}
\end{center}
   \caption{
   \textbf{Illustration of MaskConver architecture.}
   The meta architecture of MaskConver contains four components: {backbone (gray)}, {pixel decoder (pink)}, {prediction heads (light blue)}, and {mask embedding generator (green)}.
   The backbone is any commonly deployed neural network, \eg, ResNet50.
   We propose a novel ConvNeXt-UNet for the pixel decoder, which effectively captures long-range context and high-level semantics by stacking many ConvNeXt blocks at the highest level of backbone.
   We propose three prediction heads: Center Heatmap Head (for predicting center point heatmaps), Center Embedding Head (for predicting the embeddings for center points), and Mask Feature Head (for generating mask features).
   The Mask Embedding Generator first produces the class embeddings via a lookup table (Class Embedding Lookup Table module) by taking the predicted semantic classes from the top-K center points.
   The output mask embeddings are obtained by modulating the class embeddings with the center embeddings (via addition and MLP) to mitigate the center point collision between instances of different classes.
   In the end, the mask features are multiplied with the mask embeddings to generate the final binary masks.
   Unlike transformer-based methods, MaskConver only exploits convolutions without any self- or cross-attentions.
   }
\label{fig:maskconver_arch}
\end{figure*}

In this work, we revisit the pure convolution panoptic models~\cite{cheng2020panoptic,li2021fully} and propose a novel architecture for panoptic segmentation (\cref{fig:maskconver_arch}), named MaskConver, which produces segmentation masks for thing and stuff classes in a unified way.
The meta architecture of MaskConver contains four main components: backbone, pixel decoder, prediction heads, and mask embedding generator.
The backbone is a typical ImageNet~\cite{russakovsky2015imagenet} pretrained convolutional neural network (ConvNet)~\cite{lecun1998gradient}, such as ResNet~\cite{he2016deep}.
We design a novel pixel-decoder, ConvNeXt-UNet, which deploys ConvNeXt blocks~\cite{liu2022convnet} in a manner similar to UNet decoder~\cite{ronneberger2015u} but in an asymmetric way.
Particularly, in the decoder, we discover that it is critical to stack more ConvNeXt blocks at the highest level (\ie, level 5 with stride 32), which benefits the model to effectively learn context information.
The prediction heads include three predictions: Center Heatmap Prediction, Center Embedding Prediction, and Mask Feature Prediction.
The Center Heatmap prediction predicts the center point heatmaps~\cite{zhou2019objects} for \textit{both} things and stuff.
We utilize the mask centers instead of box centers to represent both things and stuff.

The Center Embedding Head generates the embeddings for the center points, while the Mask Feature Head produces the mask features.
Finally, the Mask Embedding Generator aims to generate high-quality mask embeddings by taking into account the ``instance collision'', where the center points of neighboring instances may collide, yielding the same (indistinguishable) center embeddings.
To alleviate the issue, it first produces the class embeddings (via Class Embedding Lookup Table) by taking the predicted semantic classes of the center points.
The output mask embeddings are then obtained by modulating the center embeddings with the class embeddings (via addition and MLP) to a different space conditioned on the semantic class of the instance.
Finally, the mask embeddings are multiplied with the mask features to produce segmentation masks for things and stuff in a unified way.

We evaluate the quality of MaskConver in several settings on COCO panoptic segmentation dataset~\cite{lin2014microsoft}.
On the COCO validation set, our proposed ConvNeXt-UNet pixel decoder improves the PQ of the solid pixel decoder baseline BiFPN \cite{tan2020efficientdet} by +3.1\%, while being 18\% more efficient on FLOPs.
When using the ResNet50 backbone~\cite{he2016deep}, MaskConver achieves 53.6\% and runs at 19.6 FPS on a V100 GPU.
Our model demonstrates significant gains (+9.3\%) compared to the modern convolution-based method, Panoptic FCN~\cite{li2021fully}, while also outperforming the transformer-based models like Mask2Former~\cite{cheng2022masked} (+1.7\%) and kMaX-DeepLab~\cite{yu2022kmax} (+0.6\%).
When using the MobileNet backbone~\cite{howard2019searching}, MaskConver achieves 37.2\% PQ, which is 6.4\% better than Panoptic-DeepLab~\cite{cheng2020panoptic}.
In addition, after further optimization via quantization, MaskConver runs at 30 FPS on Pixel 6 GPU, while achieving 29.7\% PQ.
\section{Related Work}
\label{sec:related_work}

Since the proposal of panoptic segmentation in~\cite{kirillov2019panoptic}, numerous efforts have emerged in this domain. Initiatives began with adaptations to existing networks, adding either a semantic~\cite{kirillov2019panopticfpn} or an instance branch~\cite{cheng2020panoptic} to state-of-the-art models These methods established a baseline using hand-crafted post-processing layers~\cite{kirillov2019panopticfpn,xiong2019upsnet,liu2019end,yang2019deeperlab} for final panoptic predictions. Following these works, researchers start to think about architectures that can solve the task in a more unified way. MaX-DeepLab~\cite{wang2021max} learns mask embeddings~\cite{jia2016dynamic,tian2020conditional,wang2020solov2} in the transformer framework for both thing and stuff classes. These mask embeddings predict a set of binary masks, and a post-processing layer~\cite{mark2021deeplab2} is used to predict the final panoptic outputs. MaskFormer~\cite{cheng2021per} proposes a similar paradigm, and shows how to use the same model for both semantic and panoptic segmentation by only modifying the post-processing logic. Panoptic-Segformer~\cite{li2022panoptic} extends Deformable DETR~\cite{zhu2021deformable} for the panoptic segmentation task.
CMT-DeepLab~\cite{yu2022cmt} reformulates the transformer cross-attention from the clustering perspective. Mask2Former~\cite{cheng2022masked} proposes masked-attention to significantly outperform MaskFormer architecture on smaller objects by masking unrelated parts of the image. kMaX-DeepLab~\cite{yu2022kmax} improves MaX-DeepLab by reformulating the cross attention layers to mimic k-means algorithm. Similarly, our MaskConver learns mask embeddings for both thing and stuff classes, but only uses fully convolutional layers~\cite{long2015fully} without any transformer blocks.

Transformers have surpassed ConvNets on several vision problems beyond panoptic segmentation~\cite{li2022mask,jain2023oneformer} including classification~\cite{dosovitskiy2021image,liu2021swin,fan2021multiscale,wang2021pyramid,dai2021coatnet,liu2022swin,yang2023moat}, detection~\cite{carion2020end,gao2021fast,li2022dn,li2022exploring,zhang2023dino}, and segmentation~\cite{zheng2021rethinking,strudel2021segmenter,wang2020axial,xu2022groupvit,dong2022cswin}. Recently, ConvNeXt~\cite{liu2022convnet}, a pure convolution-based backbone, provides competitive performance compared to transformer architectures, while preserving the simplicity and efficiency of ConvNets. ConvNeXt block adopts depthwise convolutions with a large kernel size $7\times7$, layer norm~\cite{ba2016layer}, and layer scale~\cite{touvron2021going}. In this work, we adopt the ConvNeXt block as a building block for our pixel decoder. The design is further improved by using Squeeze-and-Excitation~\cite{hu2018squeeze} layer that improves the performance with little impact on the model latency and FLOPs.

Panoptic FCN~\cite{li2021fully} employs the fully convolutional architecture~\cite{long2015fully} that predicts things and stuff classes in a unified way, through the convolutional kernel generator to predict things and stuff kernels. These kernels are used to predict binary masks. Although Panopic FCN unifies things and stuff classes towards the post-processing, the kernel generator still treats things and stuff differently (particularly, they represent things as centers, but stuff as regions).
Panoptic FCN's quality is lagging behind the recent transformer-based panoptic models.
To bridge the gap, MaskConver proposes to fully unify the architecture by only relying on things and stuff centers.
It creates a lightweight class embedding module that can break the ties when multiple centers co-exist in the same location.
MaskConver also introduces a novel pixel decoder (ConvNeXt-UNet), which provides the model with sufficient context to produce high quality centers and mask predictions.

\noindent\textbf{Efficient Panoptic Segmentation} models are less explored, since most architectures focus more on pushing the panoptic quality instead of having more efficient architectures. Panoptic-DeepLab~\cite{cheng2020panoptic,swidernet_2020} reports quality and latency numbers on a V100 GPU when using MobileNetV3 backbone~\cite{howard2019searching} on an image size of $640\times640$. Hou \etal~\cite{hou2020real} propose a single-shot panoptic segmentation model that runs in real-time on a V100 GPU.
In this work, we tailor MaskConver architecture for mobile devices (Pixel 6) through a set of architectural design choices.

Various lightweight model backbones, including MobileNet~\cite{howard2017mobilenets,sandler2018mobilenetv2,howard2019searching}, EfficientNet~\cite{tan2019efficientnet,tan2021efficientnetv2}, and ShuffleNet~\cite{zhang2018shufflenet,ma2018shufflenet}, have been proposed. They are designed for low computational power devices, like mobile CPU and GPU. A multi-hardware MobileNet (MobileNet-MH)~\cite{chu2021discovering}, discovered by neural architecture search~\cite{zoph2017neural,bender2020can} and optimized for multiple hardware~\cite{cai2020once}, achieves state-of-the-art latency and accuracy trade-off on a variety of mobile devices, and has been adopted as the backbone for the semantic segmentation task~\cite{wang2021mosaic}. To evaluate MaskConver for mobile use cases, we use MobileNet-MH since it delivers similar accuracy as MobileNetV3~\cite{howard2019searching}, while being more compatible with different mobile devices.

\noindent\textbf{Center Point Representation} is used in tasks like 2D detection, tracking, action recognition, instance segmentation and panoptic segmentation~\cite{zhou2019objects,zhou2020tracking,li2020actions,kendall2018multi,yang2019deeperlab}.
In this work, we propose to utilize the center point to model both things and stuff. Additionally, we propose to utilize the mask centers instead of box centers.

\section{Method}
\label{sec:method}

~\cref{fig:maskconver_arch} shows the meta architecture of MaskConver, containing four main components: backbone, pixel decoder,  prediction heads, and mask embedding generator.

\noindent{\bf Overview.}
The backbone is a typical convolutional neural network, such as ResNet~\cite{he2016deep} and MobileNet~\cite{howard2017mobilenets}.
A novel pixel decoder ConvNeXt-UNet is proposed to generate the image features, on top of which prediction heads are appended.
We propose three prediction heads: (1) Center Heatmap Head that predicts center point heatmaps~\cite{zhou2019objects} for \textit{both} things and stuff, (2) Center Embedding Head that predicts embeddings for the center points, and (3) Mask Feature Head that yields mask features. The Mask Embedding Generator generates the mask embeddings by taking as input both the top-K confident predicted centers (their semantic classes and coordinates) and the center embeddings. In the end, a set of binary masks are obtained by multiplying the mask features with the mask embeddings~\cite{wang2021max}. We will first explain our design motivations before detailing the proposed modules in the following subsections.

\noindent{\bf Motivations.}
Convolution-based models~\cite{cheng2020panoptic,li2021fully} lag in performance, compared to transformer-based models~\cite{cheng2022masked,yu2022kmax}.
We carefully look into the mask transformer frameworks~\cite{wang2021max,cheng2022masked,yu2022kmax}, and discover that the advantages of employing transformer blocks~\cite{vaswani2017attention} are twofold: the attention mechanism~\cite{bahdanau2014neural} effectively enriches the pixel decoder with long-range information (thus generates high quality masks) and the object queries produce thing and stuff segmentation masks in a unified way. The observation motivates us to revisit the existing convolution-based methods~\cite{cheng2020panoptic,li2021fully} by designing a better pixel decoder (~\cref{sec:convnext_unet}),  prediction heads (~\cref{sec:prediciton_heads}), and mask embedding generator (~\cref{sec:mask_embedding_generator}).

\subsection{Pixel Decoder: ConvNeXt-UNet}
\label{sec:convnext_unet}

\begin{figure}[t]
  \centering
  \includegraphics[width=0.8\linewidth]{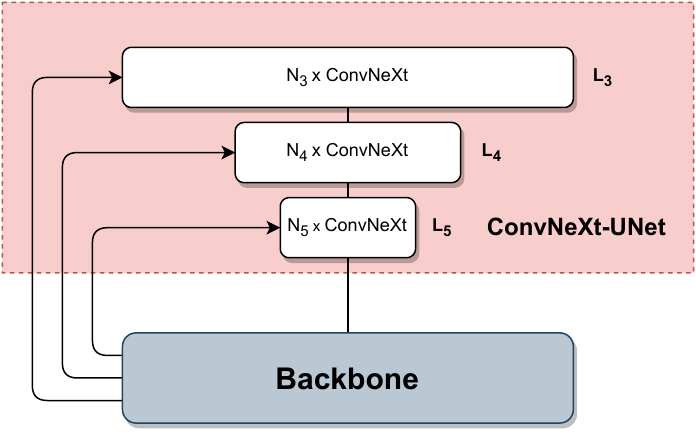}
  \caption{
  \textbf{Pixel decoder with ConvNeXt blocks}.
  We deploy the modified ConvNeXt blocks in a manner similar to UNet, but employ more blocks in the high level (\ie, level $L_5$ with stride 32).
  }
  \label{fig:context_decoder}
\end{figure}

\begin{figure}[t]
  \centering
  \includegraphics[width=.40\linewidth]{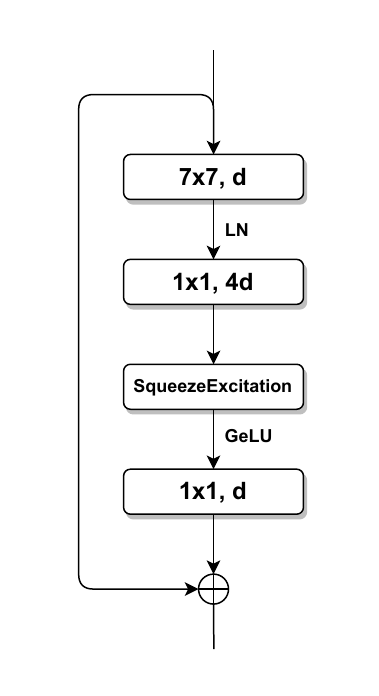}
  \caption{
  \textbf{Modified ConvNeXt block (ConvNeXt-SE)}. 
  On top of the original ConvNeXt block, we additionally include the squeeze-and-excitation operation in-between the $1\times1$ convolutions.
  }
  \label{fig:modified_convnext}
\end{figure}

To bridge the gap with transformer-based methods, we first design a novel pixel decoder ConvNeXt-UNet, as shown in ~\cref{fig:context_decoder}, consisting of the modern ConvNeXt blocks~\cite{liu2022convnet} deployed in a manner similar to UNet~\cite{ronneberger2015u} to generate image features.
Notably, ConvNeXt-UNet deploys more ConvNeXt blocks at the highest level $L_5$ of backbone (stride 32).
Thanks to the large kernel design, stacking more ConvNeXt blocks at level $L_5$ effectively captures long-range context information and high-level semantics. 
Specifically, the decoder architecture is defined by two hyper-parameters:  number of repeats, $N=[N_5, N_4, N_3]$, and channel sizes, $D=[D_5, D_4, D_3]$, determining the UNet structure from high level $L_5$ (stride 32) to low level $L_3$ (stride 8).
For example, setting $N_5 = 18$ and $D_5 = 384$ means the deployment of 18 ConvNeXt blocks with 384 channels at the level $L_5$.
Additionally, we empirically find it effective to add another Squeeze-and-Excitation~\cite{hu2018squeeze} layer in the ConvNeXt block (called ConvNeXt-SE), as shown in ~\cref{fig:modified_convnext}, which improves the model capacity at the cost of extra marginal parameters and negligible FLOPs.

\subsection{Prediction Heads}
\label{sec:prediciton_heads}

On top of the image features generated by the proposed pixel decoder, we build three prediction heads for center heatmaps, class embeddings, and mask features. Below, we first explain the structure of our prediction heads.

\noindent{\bf Light Structure of Head.}
Unlike existing methods~\cite{lin2017focal,zhou2019objects,chen2017deeplab} that commonly employ $3\times3$ convolutions in the prediction heads and introduce heavy computations on low level features (\ie, stride 8 or even stride 4 features), MaskConver, following the design principle of ConvNeXt~\cite{liu2022convnet}, adopts depthwise convolutions with a large kernel size $7\times7$ (along with layer normalization~\cite{ba2016layer} and GeLU activation function~\cite{hendrycks2016gaussian}).
This design reduces the FLOPs significantly with no degradation in the panoptic quality.

\noindent{\bf Center Heatmap Head.}
Extending object detection methods~\cite{zhou2019objects,tian2019fcos}, we propose to use center point representation for \textit{both} things and stuff. We empirically discover that mask center is a better representation than bounding box center.
The Center Heatmap Head produces a feature map with shape $H_i \times W_i \times C_{class}$, where $H_i$ and $W_i$ are the height and width of $i$-th level feature map in the feature pyramid~\cite{lin2017feature}, and $C_{class}$ is the number of semantic classes.
We will feed the top-K most confident predicted center points (their predicted semantic classes and coordinates) to the Mask Embedding Generator.

\noindent{\bf Center Embedding Head.} The Center Embedding Head generates the embeddings for center points with shape $H_i \times W_i \times C_{embd}$, where $C_{embd}$ is the channel size of embeddings. Its output is fed into the Mask Embedding Generator to gather $K$ center embeddings for the top-K most confident predicted center points (based on their coordinates).

\noindent{\bf Mask Feature Head.}
The Mask Feature Head combines the decoder features from $L_5$ to $L_3$ to create the mask features.
This is done by resizing all the decoder features to the same size (stride 4) and summing them together, before feeding to the light prediction head.
The resulting mask features have shape $H/4 \times W/4 \times C_{embd}$, where $H$ and $W$ are the height and width of input image, respectively.
The mask features, multiplied with the mask embeddings (from the Mask Embedding Generator, detailed in ~\cref{sec:mask_embedding_generator}), generate the final output: a set of $K$ binary masks.

\subsection{Mask Embedding Generator}
\label{sec:mask_embedding_generator}

The Mask Embedding Generator is one of the crucial designs in MaskConver, aiming to generate high quality mask embeddings. It takes as input both the top-K most confident predicted centers from the Center Heatmap Head (their semantic classes and coordinates) and center embeddings from the Center Embedding Head.

An na\"ive design would be the simple gathering of the K center embeddings based on the top-K center coordinates (\ie, directly use the K center embeddings as mask embeddings). 
However, it results in an inferior performance, as we observe the confusion caused by neighboring instances, especially when their centers collide, leading to exactly the same embedding vector being gathered from the output of Center Embedding Head.

Therefore, we propose to additionally exploit the \textit{class embeddings}, which learn to embed each semantic class to a vector of size $C_{embd}$. The class embeddings are used to modulate (via addition and a MLP) the center embeddings, mitigating the center collisions caused by instances of different semantic classes. Specifically, we design a ``Class Embedding Lookup Table'' module, which stores the learned embeddings for semantic classes.
For the top-K centers, we infer their most likely semantic classes, and  obtain their corresponding class embeddings from the module. We then add the obtained class embeddings and the center embeddings, and pass them to a MLP module (two fully-connected layers) to generate the final mask embeddings. We note that exploiting the class embeddings is critical to the quality of the predicted mask embeddings. It ensures that each instance will have a unique embedding vector, avoiding the problem of instance center collision.

\begin{figure}[ht]
\begin{center}
\includegraphics[width=0.6\linewidth]{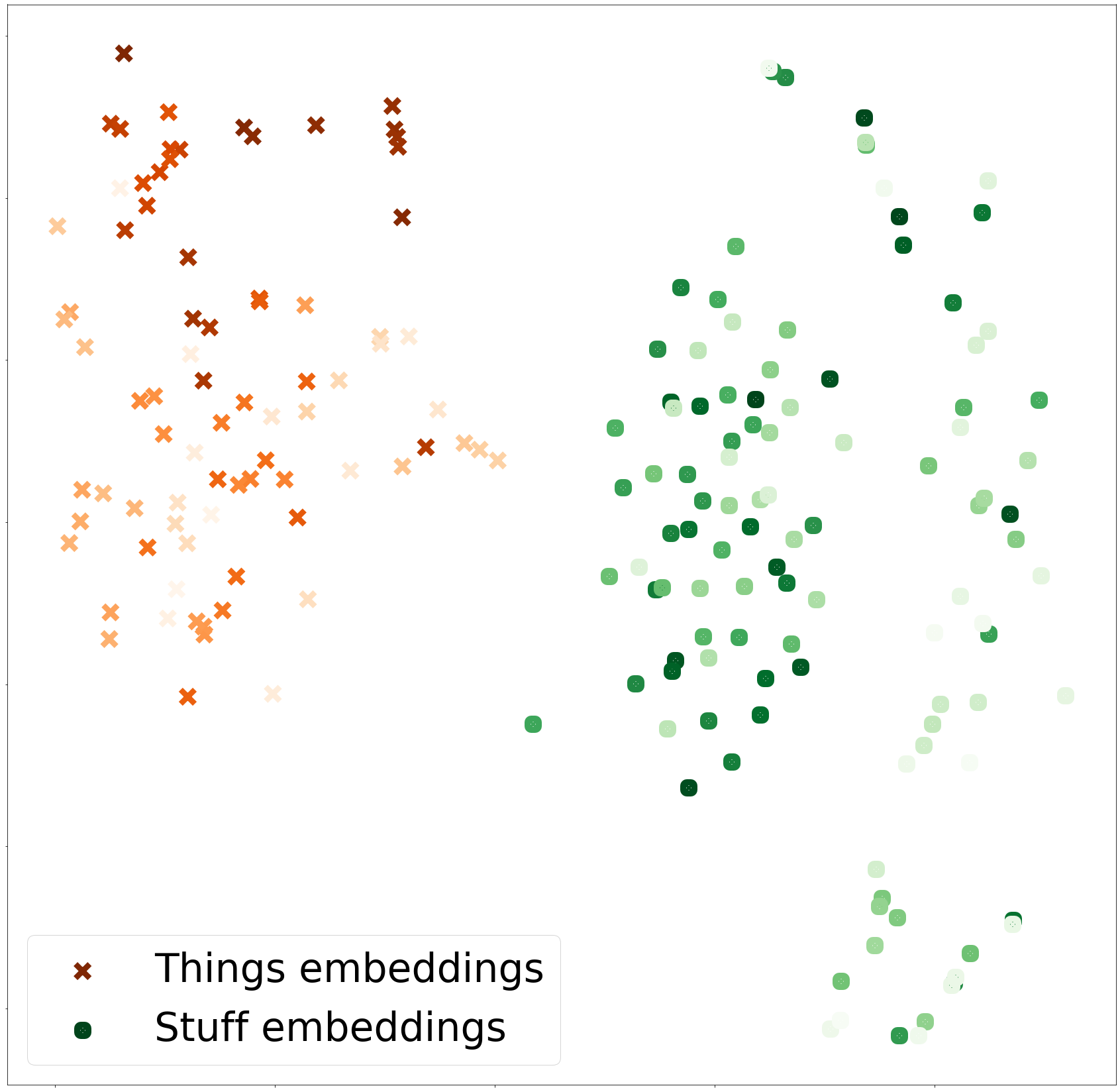}
\end{center}
   \caption{
   \textbf{tSNE~\cite{van2008visualizing} plot of learned class embeddings split by things and stuff.}
    Orange crosses are thing classes and green dots are stuff classes. 
   Our learned class embedding separates things and stuff classes automatically.
   }
\label{fig:things_class_embeddings_tsne}
\vspace{-0.4cm}
\end{figure}

~\cref{fig:things_class_embeddings_tsne} visualizes the learned class embeddings for things and stuff, using tSNE~\cite{van2008visualizing}.
As shown in the figure, there are two well-separated clusters: one for things and the other for stuff. 
As a result, the center collision between things and stuff is alleviated by adding the class embeddings and center embeddings, forming better mask embeddings to generate high quality masks.
We will show in experiments that class embeddings provide a decent performance improvements.

\section{Experimental Results}
\label{sec:results}

We evaluate the effectiveness of the proposed MaskConver on the challenging COCO dataset~\cite{lin2014microsoft}. COCO has 118,287 training images and 5,000 validation images. It has 80 thing categories and 53 stuff categories. The code and model weights will be publicly available.

\begin{table*}[ht]

\centering

\begin{tabular}{@{}l@{\ \ \ \ \ \ }c@{\ \ \ \ \ \ \ \ }c@{\ \ \ \ \ \ \ \ }c@{\ \ \ \ \ \ \ \ }c@{\ \ \ \ \ \ \ \ }c@{\ \ \ \ \ \ \ \ }c@{\ \ \ \ \ \ \ \ }c@{}}
\toprule
Architecture     & Backbone   & {Params} & {FLOPs} & {PQ} & {PQ$^{thing}$}&  {PQ$^{stuff}$} & {FPS} \\
\midrule
{Transformer Based}    \\
\hline
{\ \ \ } DETR~\cite{carion2020end}           & ResNet50~\cite{he2016deep}    & {-}       & {-}      & 43.4 & {-} & {-} & {-}                \\
{\ \ \ \ }MaskFormer~\cite{cheng2021per}     & ResNet50~\cite{he2016deep}   & 45M & 181B & 46.5 & 51.0 & 39.8 & 17.6 \\
{\ \ \ \ }K-Net~\cite{zhang2021k}     & ResNet50~\cite{he2016deep}      & {-} & {-} & 47.1 & 51.7 & 40.3 & {-} \\
{\ \ \ \ }CMT-DeepLab~\cite{yu2022cmt}    & ResNet50~\cite{he2016deep}           & {-}  & {-}  & 48.5 & {-} & {-} &  {-} \\
{\ \ \ \ }Panoptic SegFormer~\cite{li2022panoptic}   & ResNet50~\cite{he2016deep} & 51M  & 214B  & 49.6 & 54.4 & 42.4 &  7.8 \\
{\ \ \ \ }Mask2Former~\cite{cheng2022masked}   & ResNet50~\cite{he2016deep}         &  44M  & 226B & 51.9 & 57.7 & 43.0 & 8.6 \\
{\ \ \ \ }kMaX-DeepLab~\cite{yu2022kmax}     & ResNet50~\cite{he2016deep}       & 57M  & 168B    & 53.0 & 58.3 & 44.9 &  22.8  \\
\midrule
{Convolutional Based}    \\
\hline
{\ \ \ \ }Real-time Panoptic~\cite{cheng2020panoptic}$\dagger$ & ResNet50~\cite{hou2020real} & {-}      & {-}      & 37.1  & 41.0 & 31.3 & 16 \\ 
{\ \ \ \ }Panoptic FPN~\cite{kirillov2019panopticfpn} & ResNet50~\cite{he2016deep} & {-}   & {-}     & 39.0 & 45.9 & 28.7 & 17.5 \\
{\ \ \ \ }Panoptic-DeepLab~\cite{cheng2020panoptic} & Xception-71~\cite{chollet2017xception} & {-}   & 279B     & 39.7 & 43.9 & 33.2 & 7.5 \\
{\ \ \ \ }SOLO-V2~\cite{wang2020solov2} & ResNet50~\cite{he2016deep} & {-}   & {-}     & 42.1 & 49.6 & 30.7 & {-} \\
{\ \ \ \ }Unifying~\cite{li2020unifying} & ResNet50~\cite{he2016deep} & {-}   & {-}     & 43.3 & 48.6 & 35.5 & {-} \\
{\ \ \ \ }Panoptic FCN~\cite{li2021fully} & ResNet50~\cite{he2016deep}   & {-}  & {-} & 44.3  &  50.0 & 35.6 & 9.2      \\
{\ \ \ \ }MaskConver (ours)       & ResNet50~\cite{he2016deep}  & 57M  & 171B    & \textbf{53.6}   & \textbf{58.9} & \textbf{45.6} &    19.6          \\
\midrule
{Efficient Models}    \\
\hline
{\ \ \ \ }Panoptic-DeepLab~\cite{cheng2020panoptic} & MobileNetV3L~\cite{howard2019searching}  & {-} & 12.2B        & 30.0 &  {-} & {-} & {-}   \\
{\ \ \ \ }Panoptic-DeepLab~\cite{cheng2020panoptic}$\dagger$ & MobileNet-MH~\cite{chu2021discovering} & 3.9M      & 13.9B      & 30.8  & {-} & {-} & 74 \\ 
{\ \ \ \ }MaskConver (ours) & MobileNet-MH~\cite{chu2021discovering} & 3.4M      & 9.5B      & \textbf{37.2}  & 39.8 & 33.1 & 105 \\ 
{\ \ \ \ }MaskConver-256$\ddagger$ (ours) & MobileNet-MH~\cite{chu2021discovering} & 3.4M      & 1.5B      & 29.7  & 30.0 & 29.2 & \textbf{375} \\ 
\bottomrule
\end{tabular}
\caption{
\textbf{Comparison with existing models on COCO panoptic validation set.}
The FLOPs and latency are measured on a V100 GPU, using input size $800\times1200$ and $640\times640$ for ResNet50 and MobileNet backbones, respectively, following kMaX-DeepLab~\cite{yu2022kmax} and Panoptic-DeepLab~\cite{cheng2020panoptic}.
$\dagger$: Reimplemented, using the same MobileNet-MH backbone.
$\ddagger$: Method evaluated with input size $256\times256$.
}
\vspace{-3mm}
\label{tab:maskconver_vs_others}
\end{table*}

\subsection{Implementation Details}
\label{sec:implementation_details}
\noindent{\bf Training Strategy.}
The Center Heatmap Head is supervised by the Focal loss~\cite{lin2017focal}, while the final mask prediction is supervised by binary cross entropy loss and Dice loss~\cite{milletari2016v}.
The total loss is thus defined as: $Loss_{total}=\lambda_{centers} Loss_{centers} + \lambda_{bce} Loss_{bce} + \lambda_{dice} Loss_{dice}$. Across all experiments and unless stated otherwise, we fix the weighting factors as $\lambda_{centers} = 1$, $\lambda_{bce} = 10$, and $\lambda_{dice} = 10$. We use Adam weight decay optimizer (AdamW)~\cite{kingma2015adam,loshchilov2019decoupled} with learning rate of 0.001, and weight decay of 0.05. Exponential Moving Average (EMA)~\cite{polyak1992acceleration} optimizer is used with average decay of 0.99996.
We train the models for 270k iterations and batch size of 128.
The input images are resized and padded to $1280\times1280$ or $640\times640$ for ResNet50~\cite{he2016deep} or MobileNet~\cite{howard2019searching,chu2021discovering}, respectively, following~\cite{yu2022kmax,cheng2020panoptic}.
We use stronger scale jittering~\cite{ghiasi2021simple,du2021simple} with random scale of 0.05 to 2.9 (we observed overfitting if using (0.1 to 1.9)). Additionally, following~\cite{yu2022kmax}, panoptic Copy-Paste augmentation~\cite{ghiasi2021simple,kim2022tubeformer} is used.
We note that our model is built on top of TF Vision Garden~\cite{tensorflowmodelgarden2020} in TensorFlow~\cite{abadi2016tensorflow}, which does not support advanced dynamic mechanisms used in modern panoptic segmentation models~\cite{cheng2022masked}, such as deformable attention~\cite{zhu2021deformable} and uncertainty-based point supervision~\cite{kirillov2020pointrend}.

\noindent{\bf Inference Strategy.}
To predict $K$ centers, we employ a simple NMS (non-maximum suppression) layer by applying a $3 \times 3$ max-pooling to the Center Heatmap Head's output. 
The locations and corresponding semantic classes from these top-$K$ centers are subsequently used to obtain the center embeddings and class embeddings, which are combined to generate the mask embeddings and produce the final $K$ binary masks. We follow the post-processing logic of~\cite{cheng2021per} to generate panoptic segmentation outputs.
We use a score threshold of 0.2, and overlapping threshold of 0.75.

\noindent{\bf MaskConver w/ ResNet50 Backbone.}
We provide more details for MaskConver architecture, when using the ResNet50 backbone~\cite{he2016deep}.
For the proposed ConvNeXt-UNet pixel decoder, we set $N=[18, 1, 1]$ and $D=[384, 384, 384]$, \ie, stacking 18 ConvNeXt-SE blocks with 384 channels at the $L_5$ level, and only one ConvNeXt-SE block at $L_4$ and $L_3$ levels.
The Center Heatmap Head and Center Embedding Head are attached from levels $L_3$ to $L_7$, where extra strided $7\times7$ depthwise convolutions are applied to the backbone to get $L_6$ and $L_7$.

\noindent{\bf MaskConver w/ Efficient Backbone.}
We make some changes, when deloying MaskConver with efficient backbones~\cite{howard2019searching,chu2021discovering}.
The Center Heatmap Head and Center Embedding Head are only appended to a single scale, \ie, $L_3$ feature map in the feature pyramid. We experiment with MobileNetV3-Large~\cite{howard2019searching} and multi-hardware MobileNet (MobileNet-MH)~\cite{chu2021discovering} as the backbone. The ConvNeXt-UNet pixel decoder is replaced with the simpler DeepLabv3+ decoder~\cite{chen2018encoder}. For efficiency, we use hard-sigmoid~\cite{courbariaux2015binaryconnect}, since sigmoid is costly on mobile devices. The model is converted to TFLite models and benchmarked on mobile devices to obtain the latency. We also apply weight-only post-training quantization to further speed up the model by 2$\times$.

\subsection{Main Results}

In \cref{tab:maskconver_vs_others}, we compare the proposed MaskConver with other methods in three categories: convolution-based, transformer-based, and efficient models.

\noindent{\bf Convolution-Based Models.}
In the category of convolution-based Models (\ie, middle group in \cref{tab:maskconver_vs_others}), MaskConver consistently outperforms all the other convolution-based methods in terms of both performance (PQ) and speed (FPS).
Particularly, when comparing with the state-of-the-art Panoptic FCN~\cite{li2021fully}, MaskConver is +9.3\% PQ better and running 2.13 times faster.

\noindent{\bf Transformer-Based Models.}
When compared with transformer-based models (\ie, top group in \cref{tab:maskconver_vs_others}), MaskConver achieves better PQ when using similar FLOPs/parameters. In particular, MaskConver is +1.7\% better than Mask2Former~\cite{cheng2022masked}, while also being faster on a V100 GPU.
MaskConver is also +0.6\% better than kMaX-DeepLab~\cite{yu2022kmax} with a slightly higher number of flops. These results suggest that with a better designed pixel decoder, prediction heads, and mask embedding generator, MaskConver can successfully bridge the gap between transformer- and convolution-based models.

\noindent{\bf Efficient Models.}
For efficient models~\cite{howard2019searching,chu2021discovering} (\ie, bottom group in \cref{tab:maskconver_vs_others}), we compared MaskConver with Panoptic-DeepLab~\cite{cheng2020panoptic}.
We employ Panoptic-DeepLab with the same MobileNet-MH backbone~\cite{chu2021discovering} and input size $640\times640$ to have a fair comparison.
Our model with 640 input image achieves +6.4\% better PQ compared to Panoptic-DeepLab, while also being 1.42$\times$ times faster on a V100 GPU.
Furthermore, if we change the input size to $256\times256$, our MaskConver-256 achieves a similar PQ to Panoptic-DeepLab (29.7\% \vs 30.8\% PQ), while running 5.07$\times$ times faster.
Our MaskConver-256 runs real-time on Pixel 6 GPU with 33 FPS.

\begin{table*}[ht]

\centering
\begin{tabular}{l|l|cc|c}
\toprule
Pixel Decoder  & Design  & FLOPs & Params & PQ\\ 
\midrule
FPN~\cite{lin2017feature} & $L_3$-$L_7$ & 155B  & 31M &  43.1 \\
BiFPN~\cite{tan2020efficientdet} & 6 BiFPN layers, $L_3$-$L_7$& 210B  & 55M & 47.7  \\
\midrule
ConvNeXt-UNet & $N=[3, 9, 3]$, $D=[768, 384, 192]$         & 195B                       & 54M                         & 49.1                    \\
ConvNeXt-UNet & $N=[11, 1, 1]$, $D=[512, 384, 384]$& 173B               & 53M & 50.1    \\
ConvNeXt-UNet & $N=[18, 1, 1]$, $D=[384, 384, 384]$ & 171B  & 51M    & 50.4  \\
ConvNeXt-UNet & $N=[18, 1, 1]$, $D=[384, 384, 384]$ + SE~\cite{hu2018squeeze} & 171B                       & 57M                         & 50.8                    \\ 
\bottomrule
\end{tabular}
\caption{
\textbf{Effect of pixel decoder designs on COCO val set.}
Our final design of ConvNeXt-UNet pixel decoder (last row) stacks many ConvNeXt blocks at the level 5 (stride 32), effectively capturing long-range information and high-level semantics.
}
\label{table:context-decoder}
\end{table*}

\subsection{Ablation Studies}
\label{sec:ablation}

We conduct the systematic ablation studies on the MaskConver architecture, using the ResNet50 backbone.

\noindent{\bf Pixel Decoder.}
In \cref{table:context-decoder}, we ablate on the design choices of pixel decoder. We start with the popular pixel decoder choice: FPN (feature pyramid network)~\cite{lin2017feature} as our baseline, which attains the performance of 43.1\% PQ. We then use the more advanced feature pyramid architecture, BiFPN (bi-directional feature pyramid network)~\cite{tan2020efficientdet}, which improves the performance to 47.7\% PQ with a reasonable increase in both FLOPs and model parameters, serving as another solid baseline. After setting the solid baselines, we explore the structure of the proposed ConvNeXt-UNet with two hyper-parameters: number of repeats of ConvNeXt blocks, $N=[N_5, N_4, N_3]$, and channel sizes, $D=[D_5, D_4, D_3]$, which determine the UNet structure from high level $L_5$ (stride 32) to low level $L_3$ (stride 8). We first set $N=[3, 9, 3]$ and $D=[768, 384, 192]$, which corresponds to the inverting ConvNeXt-Tiny~\cite{liu2022convnet} structure. This simple design surprisingly improves over the strong baseline BiFPN by +1.4\% PQ, while also being slightly more efficient in FLOPs. Motivated by the prior works~\cite{chen2017deeplab,zhao2017pyramid} that employ the multi-scale context module at the highest level (\ie, stride 32) of backbone, we explore stacking more ConvNeXt blocks to level 5 to capture more long-range information and high-level semantics. Hence, we move most of the blocks to level 5 by setting $N=[11, 1, 1]$ and $D=[512, 383, 384]$, in order to keep the similar number of parameters. This structure further improves the performance by +1\% PQ. To further push the envelope, we explore stacking more blocks at level 5 by using $N=[18, 1, 1]$ and $D=[384, 384, 384]$, which yields additional +0.3\% improvement. Finally, employing the proposed ConvNeXt-SE block (\ie, adding another Squeeze-and-Excitation~\cite{hu2018squeeze} layer in ConvNeXt block) improves the PQ by +0.4\% with minor effect on the model FLOPs. Overall, we observe a +3.1\% improvement in PQ compared to the solid baseline BiFPN~\cite{tan2020efficientdet} with 19\% lower FLOPs, and +7.7\% improvement in PQ compared to FPN.

\noindent{\bf Light Structure of Head.}
In this study, we evaluate the effectiveness of the adopted light structure of prediction heads (in \cref{sec:prediciton_heads}).
As shown in \cref{table:efficient_heads}, the light prediction head using the $7\times7$ depthwise convolution slightly improves over the regular $3\times3$ convolution by +0.3\% PQ with a significant reduction of 75\% in FLOPs. This reduction is mainly because the convolution layers that process low-level features (\eg, stride 8 features) are very expensive. Replacing these regular convolution layers with depthwise convolution is more efficient for both accuracy and FLOPs.

\begin{table}[ht]
\centering
\begin{tabular}{l|c c | c}
\toprule
Prediction Head Structure          & FLOPs & Params & PQ \\
\midrule
$3\times3$ Convs           & 696B  & 68M    & 53.3   \\
$7\times7$ Depthwise Convs & \textbf{171B}  & \textbf{57M}    & \textbf{53.6}   \\
\bottomrule
\end{tabular}
\caption{
\textbf{Effect of using efficient heads on COCO val set.}
In the prediction heads, using $7\times7$ depthwise convolution is more efficient than $3\times3$ convolution in both accuracy and FLOPs.
}
\label{table:efficient_heads}
\end{table}

\noindent{\bf Class Embeddings.}
The proposed class embeddings (generated by Class Embedding Lookup Table in \cref{sec:mask_embedding_generator}) modulates the center embeddings (via addition and a MLP) to mitigate the instance collision. In this study, we evaluate its significance, using both ResNet50 and MobileNet-MH backbone, in \cref{table:class_embedding_impr}. As shown in the table, using the class embedding (see column `Cls-Embd') shows +1.5\% and +2.3\% improvement for ResNet50 and MobileNet-MH, respectively. These results suggest that using class embeddings helps break the tie, when two centers of different classes are present at the same location. Additionally, the improvement for MobileNet-MH is more significant, since MaskConver with MobileNet-MH uses only a single output scale ($L_3$) to predict the centers (for the purpose of efficiency), in which case we expect more center collisions and hence the class embeddings become more important.

\begin{table}[ht]
\centering
\scalebox{0.9}{
\begin{tabular}{l | c | c c c}
\toprule
Backbone & Cls-Embd  & PQ & PQ$^{thing}$ & PQ$^{stuff}$ \\
\midrule
ResNet-50 & \xmark & 52.1 & 57.2 & 44.1 \\
ResNet-50 & \cmark & \textbf{53.6} & \textbf{58.9} & \textbf{45.6} \\
\hline
MobileNet-MH & \xmark & 34.9 & 37.5 & 31.0 \\
MobileNet-MH & \cmark & \textbf{37.2} & \textbf{39.8} & \textbf{33.1} \\
\bottomrule
\end{tabular}
}
\caption{
\textbf{Significance of class embeddings on COCO val set.}
Using class embeddings (Cls-Embd) improves both things classes (PQ$^{thing}$) and stuff classes (PQ$^{stuff}$) for both ResNet50 and MobileNet-MH.
}
\label{table:class_embedding_impr}
\end{table}

\begin{table}[ht]
\centering
\begin{tabular}{l | c c | c}
\toprule
Centers               & FLOPs & Params & PQ \\
\midrule
Box Centers & 171B  & 57M    & 53.0   \\
Mask Centers           & 171B  & 57M    & \textbf{53.6}   \\
\bottomrule
\end{tabular}
\caption{
\textbf{Effect of mask centers on COCO panoptic val set.}
Using mask centers improves the PQ compared to box centers.
}
\label{table:mask_centers}
\vspace{-0.4cm}
\end{table}

\noindent{\bf Mask Centers \vs Box Centers.}
Unlike CenterNet~\cite{zhou2019objects}, MaskConver uses mask centers, instead of bounding box centers.
We ablate this design choice in \cref{table:mask_centers}, which shows +0.6\% PQ improvement, when using the mask centers.
As a result, the mask generates a better center representation than the bounding box.

\noindent{\bf Effect of Training Strategy.}
In this study, we evaluate the effect of several training techniques that are used in our framework.
As shown in \cref{table:general_improvements}, training the model longer for 270k iterations gives 0.3\% improvement over 150k iterations (used in~\cite{yu2022kmax}).
We did not observe any improvement if we train the model even longer.
Using stronger scale jittering with scale [0.05, 2.9] further improves the performance by +0.8\%.
Finally, the panoptic Copy-Paste~\cite{yu2022kmax,kim2022tubeformer} shows a significant improvement with +1.7\% PQ.
We note again that our model is built in TensorFlow~\cite{abadi2016tensorflow}, which does not support advanced dynamic mechanisms commonly used in modern panoptic segmentation models~\cite{cheng2022masked,li2022panoptic}, such as deformable attention~\cite{zhu2021deformable} and uncertainty-based point supervision~\cite{kirillov2020pointrend}.

\begin{table}[ht]
\centering
\begin{tabular}{c c c | c}
\toprule
Longer Training & Strong Aug. & Copy-Paste & PQ \\
\midrule
                &             &            &  50.8  \\
    \cmark            &             &            &  51.1 \\
    \cmark            &   \cmark          &            &  51.9  \\
    \cmark            &   \cmark          &   \cmark         &  53.6  \\
\bottomrule
\end{tabular}
\caption{
\textbf{Effect of training techniques on COCO val set.}
Our final setting employs longer training iterations (270k), strong scale augmentation ([0.05, 2.9]), and panoptic copy-paste.
}
\label{table:general_improvements}
\end{table}

\noindent{\bf Stuff Center vs Stuff Region.} We contrast Region-based and Center-based methods for representing stuff classes in MaskConver with a ResNet-50 backbone in~\cref{table:stuff_center_vs_region} on COCO val set. The center-based approach demonstrates a significant 21\% drop in latency (FPS) — a crucial advantage for real-time computer vision applications — with only a marginal dip in PQ$^{stuff}$.

\begin{table}[t]
\centering
\begin{tabular}{l | c c c}
\toprule
Stuff & Backbone &   PQ$^{stuff}$ & FPS \\
\midrule
Center & Resnet-50 &      45.6 & 19.6 \\
Region~\cite{li2021fully} & Resnet-50 &     45.8 & 15.4 \\

\bottomrule
\end{tabular}
\vspace{-0.3cm}
\caption{\textbf{Region-based vs. Center-based stuff representations using ResNet-50.} The table highlights the efficiency gains of the center-based approach in terms of latency (FPS), while showcasing a minimal compromise in PQ$^{stuff}$.}
\label{table:stuff_center_vs_region}
\end{table}

\section{Conclusion}

In this work, we have presented MaskConver, revisiting pure convolution for panoptic segmentation.
MaskConver simplifies the convolution-based panoptic models by unifying things and stuff modeling.
Specifically, MaskConver uses centers to represent both thing and stuff regions, and employs the light class embedding module to predict unique embedding vectors for multiple instances that are present at the same locations.
MaskConver also adopted the ConvNeXt-UNet pixel decoder that provides the prediction heads with long-range context and high-level semantics.
With simplified architecture and the ConvNeXt-UNet, MaskConver closes the gap with the transformer-based models on COCO dataset.
Finally, MaskConver excelled in the mobile domain, thanks to the simplicity and efficiency of convolutions.

{\small
\bibliographystyle{ieee_fullname}
\bibliography{egbib}

\begin{thebibliography}{10}\itemsep=-1pt

\bibitem{abadi2016tensorflow}
Mart{\'\i}n Abadi, Ashish Agarwal, Paul Barham, Eugene Brevdo, Zhifeng Chen, Craig Citro, Greg~S Corrado, Andy Davis, Jeffrey Dean, Matthieu Devin, et~al.
\newblock Tensorflow: Large-scale machine learning on heterogeneous distributed systems.
\newblock {\em arXiv:1603.04467}, 2016.

\bibitem{ba2016layer}
Jimmy~Lei Ba, Jamie~Ryan Kiros, and Geoffrey~E Hinton.
\newblock Layer normalization.
\newblock {\em arXiv:1607.06450}, 2016.

\bibitem{bahdanau2014neural}
Dzmitry Bahdanau, Kyunghyun Cho, and Yoshua Bengio.
\newblock Neural machine translation by jointly learning to align and translate.
\newblock In {\em ICLR}, 2015.

\bibitem{bender2020can}
Gabriel Bender, Hanxiao Liu, Bo Chen, Grace Chu, Shuyang Cheng, Pieter-Jan Kindermans, and Quoc~V Le.
\newblock Can weight sharing outperform random architecture search? an investigation with tunas.
\newblock In {\em CVPR}, 2020.

\bibitem{cai2020once}
Han Cai, Chuang Gan, Tianzhe Wang, Zhekai Zhang, and Song Han.
\newblock Once-for-all: Train one network and specialize it for efficient deployment.
\newblock In {\em ICLR}, 2020.

\bibitem{carion2020end}
Nicolas Carion, Francisco Massa, Gabriel Synnaeve, Nicolas Usunier, Alexander Kirillov, and Sergey Zagoruyko.
\newblock End-to-end object detection with transformers.
\newblock In {\em ECCV}, 2020.

\bibitem{chen2017deeplab}
Liang-Chieh Chen, George Papandreou, Iasonas Kokkinos, Kevin Murphy, and Alan~L Yuille.
\newblock Deeplab: Semantic image segmentation with deep convolutional nets, atrous convolution, and fully connected crfs.
\newblock {\em TPAMI}, 2017.

\bibitem{swidernet_2020}
Liang-Chieh Chen, Huiyu Wang, and Siyuan Qiao.
\newblock Scaling wide residual networks for panoptic segmentation.
\newblock {\em arXiv:2011.11675}, 2020.

\bibitem{chen2018encoder}
Liang-Chieh Chen, Yukun Zhu, George Papandreou, Florian Schroff, and Hartwig Adam.
\newblock Encoder-decoder with atrous separable convolution for semantic image segmentation.
\newblock In {\em ECCV)}, 2018.

\bibitem{cheng2020panoptic}
Bowen Cheng, Maxwell~D Collins, Yukun Zhu, Ting Liu, Thomas~S Huang, Hartwig Adam, and Liang-Chieh Chen.
\newblock Panoptic-deeplab: A simple, strong, and fast baseline for bottom-up panoptic segmentation.
\newblock In {\em CVPR}, 2020.

\bibitem{cheng2022masked}
Bowen Cheng, Ishan Misra, Alexander~G Schwing, Alexander Kirillov, and Rohit Girdhar.
\newblock Masked-attention mask transformer for universal image segmentation.
\newblock In {\em CVPR}, 2022.

\bibitem{cheng2021per}
Bowen Cheng, Alex Schwing, and Alexander Kirillov.
\newblock Per-pixel classification is not all you need for semantic segmentation.
\newblock {\em NeurIPS}, 2021.

\bibitem{chollet2017xception}
Fran{\c{c}}ois Chollet.
\newblock Xception: Deep learning with depthwise separable convolutions.
\newblock In {\em CVPR}, 2017.

\bibitem{chu2021discovering}
Grace Chu, Okan Arikan, Gabriel Bender, Weijun Wang, Achille Brighton, Pieter-Jan Kindermans, Hanxiao Liu, Berkin Akin, Suyog Gupta, and Andrew Howard.
\newblock Discovering multi-hardware mobile models via architecture search.
\newblock In {\em CVPR}, 2021.

\bibitem{courbariaux2015binaryconnect}
Matthieu Courbariaux, Yoshua Bengio, and Jean-Pierre David.
\newblock Binaryconnect: Training deep neural networks with binary weights during propagations.
\newblock {\em NeurIPS}, 2015.

\bibitem{dai2021coatnet}
Zihang Dai, Hanxiao Liu, Quoc~V Le, and Mingxing Tan.
\newblock Coatnet: Marrying convolution and attention for all data sizes.
\newblock {\em NeurIPS}, 34:3965--3977, 2021.

\bibitem{dong2022cswin}
Xiaoyi Dong, Jianmin Bao, Dongdong Chen, Weiming Zhang, Nenghai Yu, Lu Yuan, Dong Chen, and Baining Guo.
\newblock Cswin transformer: A general vision transformer backbone with cross-shaped windows.
\newblock In {\em CVPR}, 2022.

\bibitem{dosovitskiy2021image}
Alexey Dosovitskiy, Lucas Beyer, Alexander Kolesnikov, Dirk Weissenborn, Xiaohua Zhai, Thomas Unterthiner, Mostafa Dehghani, Matthias Minderer, Georg Heigold, Sylvain Gelly, et~al.
\newblock An image is worth 16x16 words: Transformers for image recognition at scale.
\newblock In {\em ICLR}, 2021.

\bibitem{du2021simple}
Xianzhi Du, Barret Zoph, Wei-Chih Hung, and Tsung-Yi Lin.
\newblock Simple training strategies and model scaling for object detection.
\newblock {\em arXiv:2107.00057}, 2021.

\bibitem{fan2021multiscale}
Haoqi Fan, Bo Xiong, Karttikeya Mangalam, Yanghao Li, Zhicheng Yan, Jitendra Malik, and Christoph Feichtenhofer.
\newblock Multiscale vision transformers.
\newblock In {\em ICCV}, 2021.

\bibitem{gao2021fast}
Peng Gao, Minghang Zheng, Xiaogang Wang, Jifeng Dai, and Hongsheng Li.
\newblock Fast convergence of detr with spatially modulated co-attention.
\newblock In {\em ICCV}, 2021.

\bibitem{ghiasi2021simple}
Golnaz Ghiasi, Yin Cui, Aravind Srinivas, Rui Qian, Tsung-Yi Lin, Ekin~D Cubuk, Quoc~V Le, and Barret Zoph.
\newblock Simple copy-paste is a strong data augmentation method for instance segmentation.
\newblock In {\em CVPR}, 2021.

\bibitem{hariharan2014simultaneous}
Bharath Hariharan, Pablo Arbel{\'a}ez, Ross Girshick, and Jitendra Malik.
\newblock Simultaneous detection and segmentation.
\newblock In {\em ECCV}, 2014.

\bibitem{he2016deep}
Kaiming He, Xiangyu Zhang, Shaoqing Ren, and Jian Sun.
\newblock Deep residual learning for image recognition.
\newblock In {\em CVPR}, 2016.

\bibitem{he2004multiscale}
Xuming He, Richard~S Zemel, and Miguel~{\'A} Carreira-Perpi{\~n}{\'a}n.
\newblock Multiscale conditional random fields for image labeling.
\newblock In {\em CVPR}, 2004.

\bibitem{hendrycks2016gaussian}
Dan Hendrycks and Kevin Gimpel.
\newblock Gaussian error linear units (gelus).
\newblock {\em arXiv:1606.08415}, 2016.

\bibitem{hou2020real}
Rui Hou, Jie Li, Arjun Bhargava, Allan Raventos, Vitor Guizilini, Chao Fang, Jerome Lynch, and Adrien Gaidon.
\newblock Real-time panoptic segmentation from dense detections.
\newblock In {\em CVPR}, 2020.

\bibitem{howard2019searching}
Andrew Howard, Mark Sandler, Grace Chu, Liang-Chieh Chen, Bo Chen, Mingxing Tan, Weijun Wang, Yukun Zhu, Ruoming Pang, Vijay Vasudevan, et~al.
\newblock Searching for mobilenetv3.
\newblock In {\em ICCV}, 2019.

\bibitem{howard2017mobilenets}
Andrew~G Howard, Menglong Zhu, Bo Chen, Dmitry Kalenichenko, Weijun Wang, Tobias Weyand, Marco Andreetto, and Hartwig Adam.
\newblock Mobilenets: Efficient convolutional neural networks for mobile vision applications.
\newblock {\em arXiv:1704.04861}, 2017.

\bibitem{hu2018squeeze}
Jie Hu, Li Shen, and Gang Sun.
\newblock Squeeze-and-excitation networks.
\newblock In {\em CVPR}, 2018.

\bibitem{jain2023oneformer}
Jitesh Jain, Jiachen Li, MangTik Chiu, Ali Hassani, Nikita Orlov, and Humphrey Shi.
\newblock Oneformer: One transformer to rule universal image segmentation.
\newblock In {\em CVPR}, 2023.

\bibitem{jia2016dynamic}
Xu Jia, Bert De~Brabandere, Tinne Tuytelaars, and Luc~V Gool.
\newblock Dynamic filter networks.
\newblock In {\em NeurIPS}, 2016.

\bibitem{kendall2018multi}
Alex Kendall, Yarin Gal, and Roberto Cipolla.
\newblock Multi-task learning using uncertainty to weigh losses for scene geometry and semantics.
\newblock In {\em CVPR}, 2018.

\bibitem{kim2022tubeformer}
Dahun Kim, Jun Xie, Huiyu Wang, Siyuan Qiao, Qihang Yu, Hong-Seok Kim, Hartwig Adam, In~So Kweon, and Liang-Chieh Chen.
\newblock Tubeformer-deeplab: Video mask transformer.
\newblock In {\em CVPR}, 2022.

\bibitem{kingma2015adam}
Diederik~P Kingma and Jimmy Ba.
\newblock Adam: A method for stochastic optimization.
\newblock In {\em ICLR}, 2015.

\bibitem{kirillov2019panopticfpn}
Alexander Kirillov, Ross Girshick, Kaiming He, and Piotr Doll{\'a}r.
\newblock Panoptic feature pyramid networks.
\newblock In {\em CVPR}, 2019.

\bibitem{kirillov2019panoptic}
Alexander Kirillov, Kaiming He, Ross Girshick, Carsten Rother, and Piotr Doll{\'a}r.
\newblock Panoptic segmentation.
\newblock In {\em CVPR}, 2019.

\bibitem{kirillov2020pointrend}
Alexander Kirillov, Yuxin Wu, Kaiming He, and Ross Girshick.
\newblock Pointrend: Image segmentation as rendering.
\newblock In {\em CVPR}, 2020.

\bibitem{lecun1998gradient}
Yann LeCun, L{\'e}on Bottou, Yoshua Bengio, and Patrick Haffner.
\newblock Gradient-based learning applied to document recognition.
\newblock {\em Proceedings of the IEEE}, 86(11):2278--2324, 1998.

\bibitem{li2022dn}
Feng Li, Hao Zhang, Shilong Liu, Jian Guo, Lionel~M Ni, and Lei Zhang.
\newblock Dn-detr: Accelerate detr training by introducing query denoising.
\newblock In {\em CVPR}, 2022.

\bibitem{li2022mask}
Feng Li, Hao Zhang, Shilong Liu, Lei Zhang, Lionel~M Ni, Heung-Yeung Shum, et~al.
\newblock Mask dino: Towards a unified transformer-based framework for object detection and segmentation.
\newblock {\em arXiv:2206.02777}, 2022.

\bibitem{li2020unifying}
Qizhu Li, Xiaojuan Qi, and Philip~HS Torr.
\newblock Unifying training and inference for panoptic segmentation.
\newblock In {\em CVPR}, 2020.

\bibitem{li2022exploring}
Yanghao Li, Hanzi Mao, Ross Girshick, and Kaiming He.
\newblock Exploring plain vision transformer backbones for object detection.
\newblock In {\em ECCV}, 2022.

\bibitem{li2020actions}
Yixuan Li, Zixu Wang, Limin Wang, and Gangshan Wu.
\newblock Actions as moving points.
\newblock In {\em ECCV}, 2020.

\bibitem{li2021fully}
Yanwei Li, Hengshuang Zhao, Xiaojuan Qi, Liwei Wang, Zeming Li, Jian Sun, and Jiaya Jia.
\newblock Fully convolutional networks for panoptic segmentation.
\newblock In {\em CVPR}, 2021.

\bibitem{li2022panoptic}
Zhiqi Li, Wenhai Wang, Enze Xie, Zhiding Yu, Anima Anandkumar, Jose~M Alvarez, Ping Luo, and Tong Lu.
\newblock Panoptic segformer: Delving deeper into panoptic segmentation with transformers.
\newblock In {\em CVPR}, 2022.

\bibitem{lin2017feature}
Tsung-Yi Lin, Piotr Doll{\'a}r, Ross Girshick, Kaiming He, Bharath Hariharan, and Serge Belongie.
\newblock Feature pyramid networks for object detection.
\newblock In {\em CVPR}, 2017.

\bibitem{lin2017focal}
Tsung-Yi Lin, Priya Goyal, Ross Girshick, Kaiming He, and Piotr Doll{\'a}r.
\newblock Focal loss for dense object detection.
\newblock In {\em ICCV}, 2017.

\bibitem{lin2014microsoft}
Tsung-Yi Lin, Michael Maire, Serge Belongie, James Hays, Pietro Perona, Deva Ramanan, Piotr Doll{\'a}r, and C~Lawrence Zitnick.
\newblock Microsoft coco: Common objects in context.
\newblock In {\em ECCV}, 2014.

\bibitem{liu2019end}
Huanyu Liu, Chao Peng, Changqian Yu, Jingbo Wang, Xu Liu, Gang Yu, and Wei Jiang.
\newblock An end-to-end network for panoptic segmentation.
\newblock In {\em CVPR}, 2019.

\bibitem{liu2022swin}
Ze Liu, Han Hu, Yutong Lin, Zhuliang Yao, Zhenda Xie, Yixuan Wei, Jia Ning, Yue Cao, Zheng Zhang, Li Dong, et~al.
\newblock Swin transformer v2: Scaling up capacity and resolution.
\newblock In {\em CVPR}, 2022.

\bibitem{liu2021swin}
Ze Liu, Yutong Lin, Yue Cao, Han Hu, Yixuan Wei, Zheng Zhang, Stephen Lin, and Baining Guo.
\newblock Swin transformer: Hierarchical vision transformer using shifted windows.
\newblock In {\em ICCV}, 2021.

\bibitem{liu2022convnet}
Zhuang Liu, Hanzi Mao, Chao-Yuan Wu, Christoph Feichtenhofer, Trevor Darrell, and Saining Xie.
\newblock A convnet for the 2020s.
\newblock In {\em CVPR}, 2022.

\bibitem{long2015fully}
Jonathan Long, Evan Shelhamer, and Trevor Darrell.
\newblock Fully convolutional networks for semantic segmentation.
\newblock In {\em CVPR}, 2015.

\bibitem{loshchilov2019decoupled}
Ilya Loshchilov and Frank Hutter.
\newblock Decoupled weight decay regularization.
\newblock In {\em ICLR}, 2019.

\bibitem{ma2018shufflenet}
Ningning Ma, Xiangyu Zhang, Hai-Tao Zheng, and Jian Sun.
\newblock Shufflenet v2: Practical guidelines for efficient cnn architecture design.
\newblock In {\em ECCV}, 2018.

\bibitem{milletari2016v}
Fausto Milletari, Nassir Navab, and Seyed-Ahmad Ahmadi.
\newblock V-net: Fully convolutional neural networks for volumetric medical image segmentation.
\newblock In {\em 3DV}, 2016.

\bibitem{polyak1992acceleration}
Boris~T Polyak and Anatoli~B Juditsky.
\newblock Acceleration of stochastic approximation by averaging.
\newblock {\em SIAM journal on control and optimization}, 30(4):838--855, 1992.

\bibitem{ronneberger2015u}
Olaf Ronneberger, Philipp Fischer, and Thomas Brox.
\newblock U-net: Convolutional networks for biomedical image segmentation.
\newblock In {\em MICCAI}, 2015.

\bibitem{russakovsky2015imagenet}
Olga Russakovsky, Jia Deng, Hao Su, Jonathan Krause, Sanjeev Satheesh, Sean Ma, Zhiheng Huang, Andrej Karpathy, Aditya Khosla, Michael~S. Bernstein, Alexander~C. Berg, and Li Fei-Fei.
\newblock Imagenet large scale visual recognition challenge.
\newblock {\em IJCV}, 115:211--252, 2015.

\bibitem{sandler2018mobilenetv2}
Mark Sandler, Andrew Howard, Menglong Zhu, Andrey Zhmoginov, and Liang-Chieh Chen.
\newblock Mobilenetv2: Inverted residuals and linear bottlenecks.
\newblock In {\em CVPR}, 2018.

\bibitem{strudel2021segmenter}
Robin Strudel, Ricardo Garcia, Ivan Laptev, and Cordelia Schmid.
\newblock Segmenter: Transformer for semantic segmentation.
\newblock In {\em ICCV}, 2021.

\bibitem{tan2019efficientnet}
Mingxing Tan and Quoc Le.
\newblock Efficientnet: Rethinking model scaling for convolutional neural networks.
\newblock In {\em ICML}. PMLR, 2019.

\bibitem{tan2021efficientnetv2}
Mingxing Tan and Quoc Le.
\newblock Efficientnetv2: Smaller models and faster training.
\newblock In {\em ICML}, 2021.

\bibitem{tan2020efficientdet}
Mingxing Tan, Ruoming Pang, and Quoc~V Le.
\newblock Efficientdet: Scalable and efficient object detection.
\newblock In {\em CVPR}, 2020.

\bibitem{tian2020conditional}
Zhi Tian, Chunhua Shen, and Hao Chen.
\newblock Conditional convolutions for instance segmentation.
\newblock In {\em ECCV}, 2020.

\bibitem{tian2019fcos}
Zhi Tian, Chunhua Shen, Hao Chen, and Tong He.
\newblock Fcos: Fully convolutional one-stage object detection.
\newblock In {\em ICCV}, 2019.

\bibitem{touvron2021going}
Hugo Touvron, Matthieu Cord, Alexandre Sablayrolles, Gabriel Synnaeve, and Herv{\'e} J{\'e}gou.
\newblock Going deeper with image transformers.
\newblock In {\em CVPR}, 2021.

\bibitem{van2008visualizing}
Laurens Van~der Maaten and Geoffrey Hinton.
\newblock Visualizing data using t-sne.
\newblock {\em JMLR}, 2008.

\bibitem{vaswani2017attention}
Ashish Vaswani, Noam Shazeer, Niki Parmar, Jakob Uszkoreit, Llion Jones, Aidan~N Gomez, {\L}ukasz Kaiser, and Illia Polosukhin.
\newblock Attention is all you need.
\newblock In {\em NeurIPS}, 2017.

\bibitem{wang2021max}
Huiyu Wang, Yukun Zhu, Hartwig Adam, Alan Yuille, and Liang-Chieh Chen.
\newblock Max-deeplab: End-to-end panoptic segmentation with mask transformers.
\newblock In {\em CVPR}, 2021.

\bibitem{wang2020axial}
Huiyu Wang, Yukun Zhu, Bradley Green, Hartwig Adam, Alan Yuille, and Liang-Chieh Chen.
\newblock {Axial-DeepLab: Stand-Alone Axial-Attention for Panoptic Segmentation}.
\newblock In {\em ECCV}, 2020.

\bibitem{wang2021mosaic}
Weijun Wang and Andrew Howard.
\newblock Mosaic: Mobile segmentation via decoding aggregated information and encoded context.
\newblock {\em arXiv:2112.11623}, 2021.

\bibitem{wang2021pyramid}
Wenhai Wang, Enze Xie, Xiang Li, Deng-Ping Fan, Kaitao Song, Ding Liang, Tong Lu, Ping Luo, and Ling Shao.
\newblock Pyramid vision transformer: A versatile backbone for dense prediction without convolutions.
\newblock In {\em ICCV}, 2021.

\bibitem{wang2020solov2}
Xinlong Wang, Rufeng Zhang, Tao Kong, Lei Li, and Chunhua Shen.
\newblock Solov2: Dynamic and fast instance segmentation.
\newblock {\em NeurIPS}, 33:17721--17732, 2020.

\bibitem{mark2021deeplab2}
Mark Weber, Huiyu Wang, Siyuan Qiao, Jun Xie, Maxwell~D. Collins, Yukun Zhu, Liangzhe Yuan, Dahun Kim, Qihang Yu, Daniel Cremers, Laura Leal-Taixe, Alan~L. Yuille, Florian Schroff, Hartwig Adam, and Liang-Chieh Chen.
\newblock {DeepLab2: A TensorFlow Library for Deep Labeling}.
\newblock {\em arXiv: 2106.09748}, 2021.

\bibitem{xiong2019upsnet}
Yuwen Xiong, Renjie Liao, Hengshuang Zhao, Rui Hu, Min Bai, Ersin Yumer, and Raquel Urtasun.
\newblock Upsnet: A unified panoptic segmentation network.
\newblock In {\em CVPR}, 2019.

\bibitem{xu2022groupvit}
Jiarui Xu, Shalini De~Mello, Sifei Liu, Wonmin Byeon, Thomas Breuel, Jan Kautz, and Xiaolong Wang.
\newblock Groupvit: Semantic segmentation emerges from text supervision.
\newblock In {\em CVPR}, 2022.

\bibitem{yang2023moat}
Chenglin Yang, Siyuan Qiao, Qihang Yu, Xiaoding Yuan, Yukun Zhu, Alan Yuille, Hartwig Adam, and Liang-Chieh Chen.
\newblock Moat: Alternating mobile convolution and attention brings strong vision models.
\newblock In {\em ICLR}, 2023.

\bibitem{yang2019deeperlab}
Tien-Ju Yang, Maxwell~D Collins, Yukun Zhu, Jyh-Jing Hwang, Ting Liu, Xiao Zhang, Vivienne Sze, George Papandreou, and Liang-Chieh Chen.
\newblock Deeperlab: Single-shot image parser.
\newblock {\em arXiv:1902.05093}, 2019.

\bibitem{tensorflowmodelgarden2020}
Hongkun Yu, Chen Chen, Xianzhi Du, Yeqing Li, Abdullah Rashwan, Le Hou, Pengchong Jin, Fan Yang, Frederick Liu, Jaeyoun Kim, and Jing Li.
\newblock Tensorflow model garden.
\newblock \url{https://github.com/tensorflow/models}, 2020.

\bibitem{yu2022cmt}
Qihang Yu, Huiyu Wang, Dahun Kim, Siyuan Qiao, Maxwell Collins, Yukun Zhu, Hartwig Adam, Alan Yuille, and Liang-Chieh Chen.
\newblock Cmt-deeplab: Clustering mask transformers for panoptic segmentation.
\newblock In {\em CVPR}, 2022.

\bibitem{yu2022kmax}
Qihang Yu, Huiyu Wang, Siyuan Qiao, Maxwell Collins, Yukun Zhu, Hartwig Adam, Alan Yuille, and Liang-Chieh Chen.
\newblock k-means mask transformer.
\newblock In {\em ECCV}, 2022.

\bibitem{zhang2023dino}
Hao Zhang, Feng Li, Shilong Liu, Lei Zhang, Hang Su, Jun Zhu, Lionel Ni, and Heung-Yeung Shum.
\newblock {DINO}: {DETR} with improved denoising anchor boxes for end-to-end object detection.
\newblock In {\em ICLR}, 2023.

\bibitem{zhang2021k}
Wenwei Zhang, Jiangmiao Pang, Kai Chen, and Chen~Change Loy.
\newblock K-net: Towards unified image segmentation.
\newblock {\em NeurIPS}, 34:10326--10338, 2021.

\bibitem{zhang2018shufflenet}
Xiangyu Zhang, Xinyu Zhou, Mengxiao Lin, and Jian Sun.
\newblock Shufflenet: An extremely efficient convolutional neural network for mobile devices.
\newblock In {\em CVPR}, 2018.

\bibitem{zhao2017pyramid}
Hengshuang Zhao, Jianping Shi, Xiaojuan Qi, Xiaogang Wang, and Jiaya Jia.
\newblock Pyramid scene parsing network.
\newblock In {\em CVPR}, 2017.

\bibitem{zheng2021rethinking}
Sixiao Zheng, Jiachen Lu, Hengshuang Zhao, Xiatian Zhu, Zekun Luo, Yabiao Wang, Yanwei Fu, Jianfeng Feng, Tao Xiang, Philip~HS Torr, et~al.
\newblock Rethinking semantic segmentation from a sequence-to-sequence perspective with transformers.
\newblock In {\em CVPR}, 2021.

\bibitem{zhou2020tracking}
Xingyi Zhou, Vladlen Koltun, and Philipp Kr{\"a}henb{\"u}hl.
\newblock Tracking objects as points.
\newblock In {\em ECCV}, 2020.

\bibitem{zhou2019objects}
Xingyi Zhou, Dequan Wang, and Philipp Kr{\"a}henb{\"u}hl.
\newblock Objects as points.
\newblock {\em arXiv:1904.07850}, 2019.

\bibitem{zhu2021deformable}
Xizhou Zhu, Weijie Su, Lewei Lu, Bin Li, Xiaogang Wang, and Jifeng Dai.
\newblock Deformable detr: Deformable transformers for end-to-end object detection.
\newblock In {\em ICLR}, 2021.

\bibitem{zoph2017neural}
Barret Zoph and Quoc~V Le.
\newblock Neural architecture search with reinforcement learning.
\newblock In {\em ICLR}, 2017.

\end{thebibliography}
}

\end{document}